\documentclass[10pt,twocolumn,letterpaper]{article}
\usepackage{cvpr}
\usepackage{times}
\usepackage{epsfig}
\usepackage{graphicx}
\usepackage{amsmath}
\usepackage{amssymb}
\usepackage{mathtools}
\usepackage{subcaption}
\usepackage{booktabs}
\usepackage{array}
\usepackage{xcolor}

\newcolumntype{L}{>{\centering\arraybackslash}m{3cm}}

\def\ie{\emph{i.e}\onedot}

\usepackage[pagebackref=true,breaklinks=true,letterpaper=true,colorlinks,bookmarks=false]{hyperref}

\cvprfinalcopy 

\def\cvprPaperID{09651} 

\ifcvprfinal\fi
\begin{document}

\title{Just Go with the Flow: Self-Supervised Scene Flow Estimation}

\author{Himangi Mittal \\
Carnegie Mellon University\\
{\tt\small hmittal@andrew.cmu.edu}
\and
Brian Okorn \\
Carnegie Mellon University\\
{\tt\small bokorn@andrew.cmu.edu}
\and
David Held\\
Carnegie Mellon University\\
{\tt\small dheld@andrew.cmu.edu}
}

\maketitle
\thispagestyle{empty}

\begin{abstract}
When interacting with highly dynamic environments, scene flow allows autonomous systems to reason about the non-rigid motion of multiple independent objects. This is of particular interest in the field of autonomous driving, in which many cars, people, bicycles, and other objects need to be accurately tracked. Current state-of-the-art methods require annotated scene flow data from autonomous driving scenes to train scene flow networks with supervised learning. As an alternative, we present a method of training scene flow that uses two self-supervised losses, based on nearest neighbors and cycle consistency.  These self-supervised losses allow us to train our method on large unlabeled autonomous driving datasets; the resulting method matches current state-of-the-art supervised performance using no real world annotations and exceeds state-of-the-art performance when combining our self-supervised approach with supervised learning on a smaller labeled dataset. 



\end{abstract}
\section{Introduction}

For an autonomous vehicle, understanding the dynamics of the surrounding environment is critical to ensure safe planning and navigation. It is essential for a self-driving vehicle to be able to perceive the actions of various entities around it, such as other vehicles, pedestrians, and cyclists. In the context of data recorded as 3D point clouds, a motion can be estimated for each 3D point; this is known as scene flow, which refers to the 3D velocity of each 3D point in a scene. 
Its 2D analog, optical flow, is the projection of scene flow onto the image plane of a camera. An alternative to scene flow estimation is to use 3D object detection for object-level tracking and to assume that all points within a bounding box have the same rigid motion. However, in such a pipeline, errors in object detection can lead to errors in tracking. In contrast, scene flow methods can avoid such errors by directly estimating the motion of each 3D point.



Recent state-of-the-art methods learn to estimate the scene flow from 3D point clouds~\cite{liu2019flownet3d,gu2019hplflownet,wang2018deep, pointpwcnet}. However, these methods are fully supervised and require annotated datasets for training.  Such annotations are costly to obtain as they require labeling the motion for every point in a scene. To compensate for the lack of real world data, learning-based methods for scene flow have been trained primarily on synthetic datasets and fine tuned on real world data. This requirement of labeled training data limits the effectiveness of such methods in real world settings.

\begin{figure}[t]
 	\centering
 	\includegraphics[trim={5 10 25 0}, clip,width=0.45\textwidth]{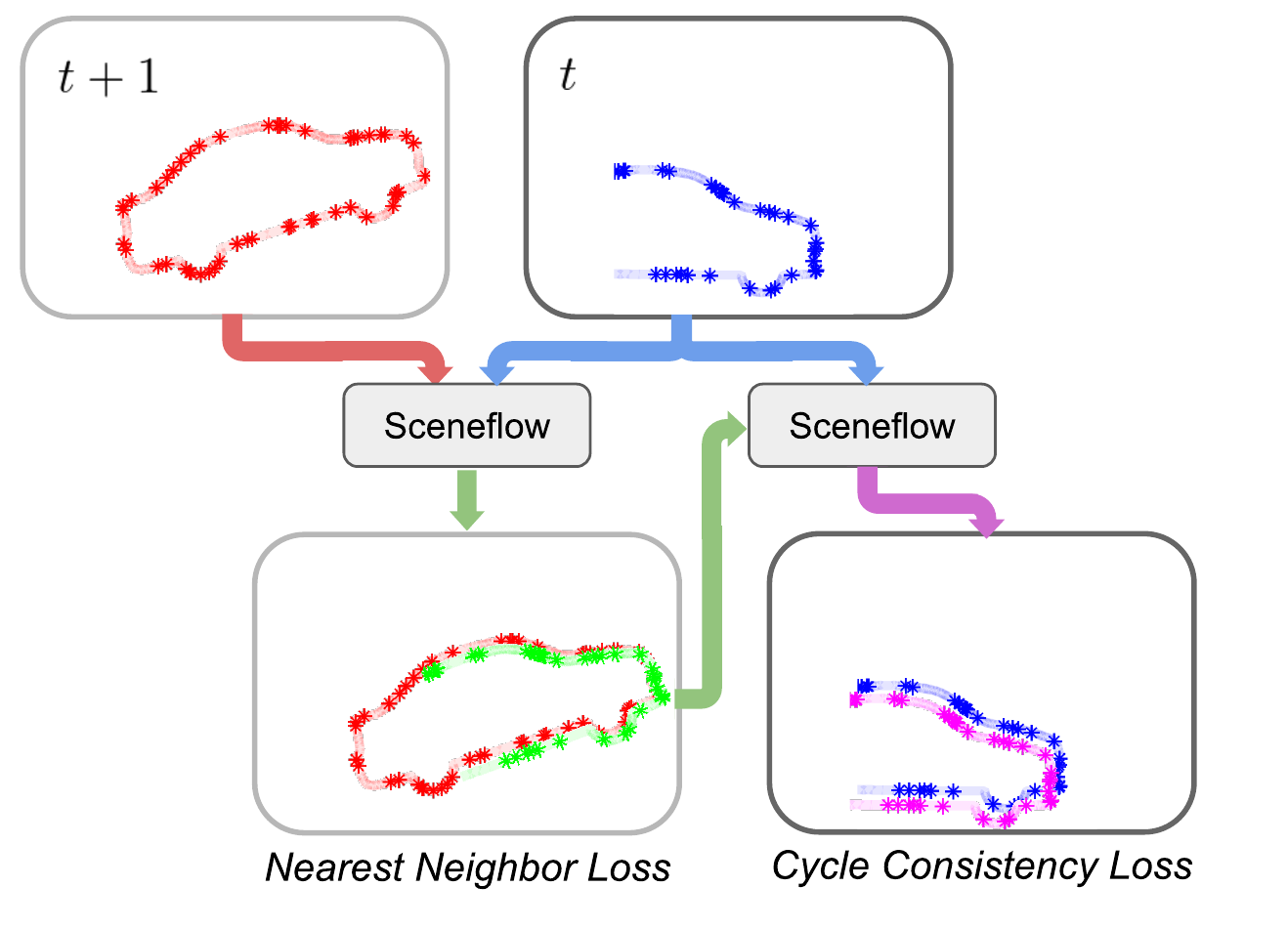}
 	\caption{We use two self-supervised losses to learn scene flow on large unlabeled datasets.  The ``nearest neighbor loss" penalizes the distance between the predicted point cloud (\textcolor{green}{green}) and each predicted point's nearest neighbor in the second point cloud (\textcolor{red}{red}).  To avoid degenerate solutions, we also estimate the flow between these predicted points (\textcolor{green}{green}) in the reverse direction back to the original point cloud (\textcolor{blue}{blue}) to form a cycle. The new predicted points from the cycle (\textcolor{magenta}{purple}) should align with the original points (\textcolor{blue}{blue}); the distance between these two set of points forms our second self-supervised loss: ``cycle consistency."}
 	\label{fig:cycle}
  	\vspace{-1em}
\end{figure} 







To overcome this limitation, we propose a self-supervised method for scene flow estimation. Using a combination of two self-supervised losses, we are able to mimic the supervision created by human annotation. Specifically, we use a cycle consistency loss, which  ensures that the scene flow produced is consistent in time (\ie we ensure that a temporal cycle ends where it started). We also use a nearest neighbor loss; due to the unavailability of scene flow annotations, we consider the nearest point to the predicted translated point, in the temporally next point cloud, as the pseudo-ground truth association.  Intuitively, the nearest neighbor loss pushes one point cloud to flow toward occupied regions of the next point cloud. We show that this combination of losses can be used to train a scene flow network over large-scale, unannotated datasets containing sequential point cloud data. An overview of our method can be found in Figure~\ref{fig:cycle}.

We test our self-supervised training approach using the neural network architecture of a state-of-the-art scene flow method~\cite{liu2019flownet3d}.  The self-supervision allows us to train this network on large-scale, unlabeled autonomous driving datasets. Our method matches the current state-of-the-art performance when no real world annotations are given. Moreover, our method exceeds the performance of state-of-the-art scene flow estimation methods when combined with supervised learning on a smaller labeled dataset. 

\section{Related Work}
\paragraph*{Scene Flow} Vedula \etal~\cite{vedula1999three} first introduced the task of scene flow estimation. They propose a linear algorithm to compute it from optical flow. Other works involve joint optimization of camera extrinsics and depth estimates for stereo scene flow~\cite{valgaerts2010joint}, use of particle filters~\cite{hadfield2011kinecting}, local rigid motion priors~\cite{vogel20113d,vogel20153d,vogel2013piecewise, menze2015object}, and smoothness-based regularization~\cite{basha2013multi}. 

\vspace{-1em}
\paragraph*{Deep Scene Flow} 
State-of-the-art scene flow methods today use deep learning to improve performance.  
FlowNet3D~\cite{liu2019flownet3d} builds on PointNet++~\cite{qi2017pointnet++, qi2017pointnet} to estimate scene flow directly from a pair of point clouds.
Gu \etal ~\cite{gu2019hplflownet} produced similar results using a permutohedral lattice to encode the point cloud data in a sparse, structured manner.  The above approaches compute scene flow directly from 3D point clouds~\cite{liu2019flownet3d,gu2019hplflownet}. Methods involving voxelizations with object-centeric rigid body assumptions~\cite{Behl_2019_CVPR}, range images~\cite{vaquero2018deep}, and non-grid structured data~\cite{wang2018deep} have also been used for scene flow estimation. All of the above methods were trained either with synthetic data~\cite{mayer2016large} or with a small amount of annotated real-world data~\cite{kitti} (or both).  Our self-supervised losses enable training on large unlabeled datasets, leading to large improvements in performance.






\vspace{-1em}
\paragraph*{Self-Supervised Learning} Wang \etal ~\cite{wang2019learning} used self-supervised learning for 2D tracking on video. They propose a tracker which takes a patch of an image at time $t$ and the entire image at time $t-1$ to track the image patch in the previous frame. They define a self-supervised loss by tracking the patch forward and backward in time to form a cycle while penalizing the errors through cycle consistency and feature similarity. We take inspiration from this work for our self-supervised flow estimation on point clouds. Other works including self-supervisory signals are image frame ordering~\cite{misra2016shuffle}, feature similarity over time~\cite{wang2015unsupervised} for images, and clustering and reconstruction~\cite{hassani2019ICCV} from point clouds. While these can potentially be used for representation learning from 3D data, they cannot be directly used for scene flow estimation. Concurrent to our work, Wu \etal~\cite{pointpwcnet} showed that Chamfer distance, smoothness constraints, and Laplacian regularization can be used to train scene flow in a self-supervised manner.



\section{Method}

\subsection{Problem Definition}
For the task of scene flow estimation, we have a temporal sequence of point clouds: point cloud $\mathcal{X}$ as the point cloud captured at time $t$ and point cloud $\mathcal{Y}$ captured at time $t+1$. There is no structure enforced on these point clouds and they can be recorded directly from a LIDAR sensor or estimated through a stereo algorithm. 
Each point $p_i = \{x_i,f_i\}$ in point cloud $\mathcal{X}$ contains the Cartesian position of the point, $x_i \in \mathbb{R}^3$, as well as any additional information which the sensor produces, such as color, intensity, normals, etc, represented by $f_i \in \mathbb{R}^c$.

The scene flow, $\mathcal{D} = \{d_i\}^N, d_i \in \mathbb{R}^3$ between these two point clouds describes the movement of each point $x_i$ in point cloud $\mathcal{X}$ to its corresponding position $x_i'$ in the scene described by point cloud $\mathcal{Y}$, such that $x_i' = x_i + d_i$,  and $N$ is the size of point cloud $\mathcal{X}$. Scene flow is defined such that $x_i$ and $x'_i$ represent the same 3D point of an object moved in time. Unlike optical flow estimation, the exact 3D position of $x_i'$ may not necessarily coincide with a point in the point cloud $\mathcal{Y}$, due to the sparsity of the point cloud. Additionally, the sizes of $\mathcal{X}$ and $\mathcal{Y}$ may be different.

\vspace{-1em}
\paragraph*{Supervised Loss}
The true error associated with our task is the difference between the estimated flow $g(\mathcal{X}, \mathcal{Y}) = \mathcal{\hat{D}} = \{\hat{d}_i\}^N$ and the ground truth flow $\mathcal{D}^* = \{d^*_i\}^N$, 
\begin{align}
\mathcal{L}_{gt} = \frac{1}{N} \sum_i^N \|d_i^*-\hat{d}_i\|^2.   
\label{equ:gt}
\end{align}
The loss in Equation~\ref{equ:gt} is useful because it is mathematically equivalent to the end point error, which we use as our evaluation metric. However, computing this loss requires annotated ground truth flow $d^*_i$. This type of annotation is easy to calculate in synthetic data~\cite{mayer2016large}, but requires expensive human labeling for real world datasets. As such, only a small amount of annotated scene flow datasets are available ~\cite{menze2015joint, menze2018object}. While training on  purely synthetic data is possible, large improvements can often be obtained by training on real data from the domain of the target application.  For example, Lui \etal~\cite{liu2019flownet3d} showed an 18\% relative improvement after fine-tuning on a small amount of annotated real world data.  This result motivates our work to use self-supervised training to train on large unlabeled datasets. 

\begin{figure}[t]
 	\centering
 	\begin{subfigure}{0.23\textwidth}
 	\includegraphics[trim={70 250 70 250}, clip, width=\linewidth]{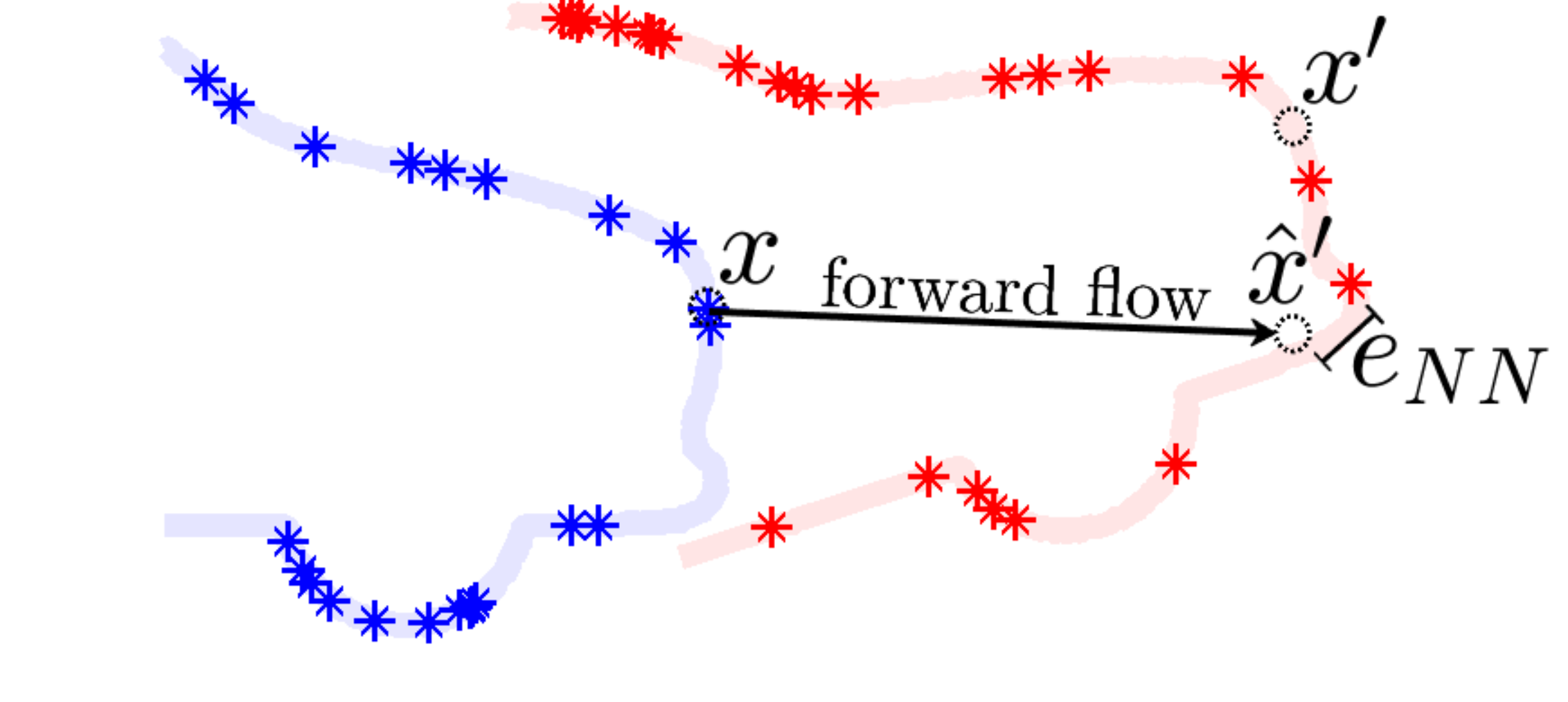}
 	\caption{Nearest Neighbor Loss}
 	\end{subfigure}
  	\begin{subfigure}{0.23\textwidth}
 	\includegraphics[trim={70 250 70 250}, clip, width=\linewidth]{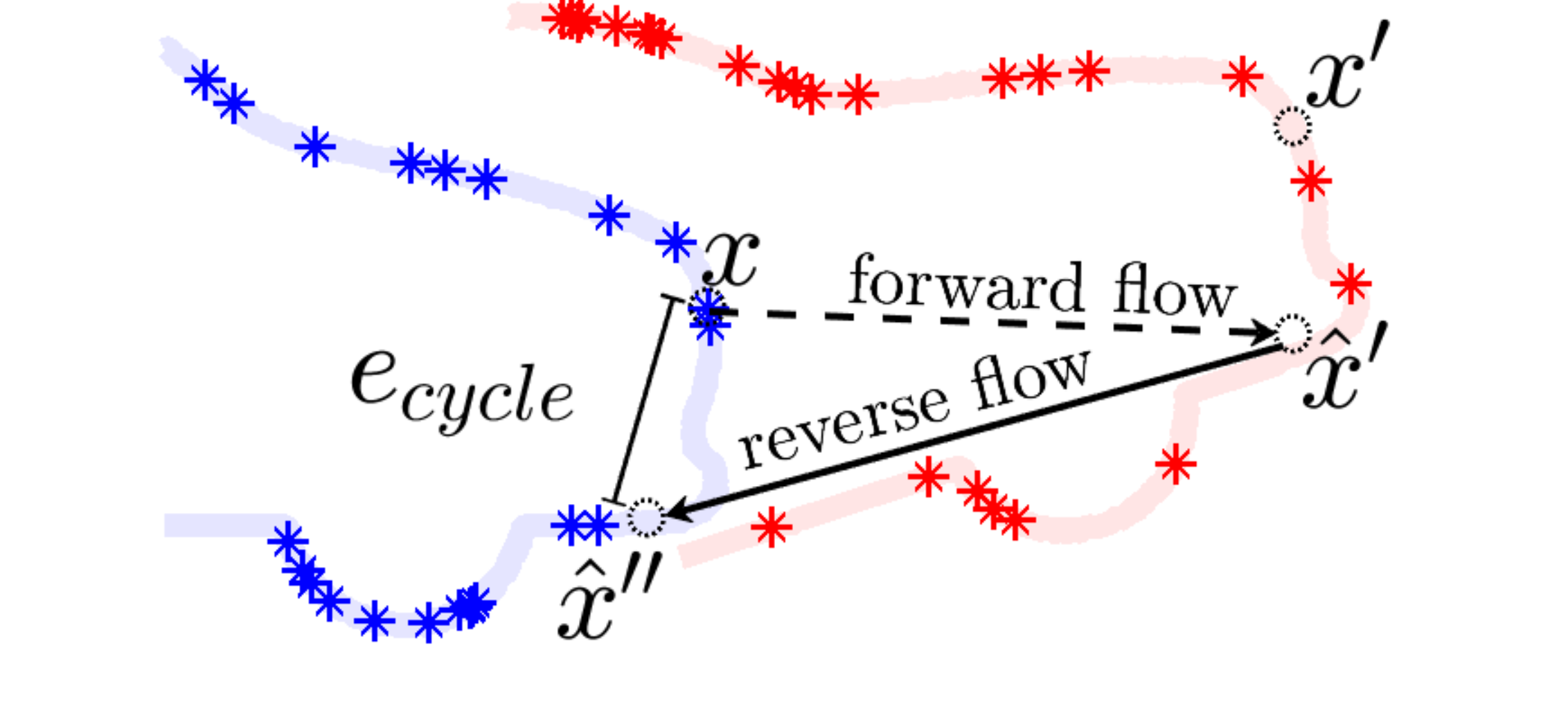}
 	\caption{Cycle Consistency Loss}
 	\end{subfigure}
 	\caption{Example of our self-supervised losses between consecutive point clouds $\mathcal{X}$ (\textcolor{blue}{blue}) and $\mathcal{Y}$ (\textcolor{red}{red}). We consider the point $x$ whose ground truth projected point $x'$ is not known during training time. $(a)$ Nearest Neighbor Loss is computed between the projected point $\hat{x}'$, predicted by the forward flow, and the closest point in $\mathcal{Y}$.  $(b)$ The Cycle Consistency Loss tracks this transformed point back onto its original frame, as point $\hat{x}''$, using the reverse flow, and computes the distance to its original position ${x}$.}
 	\label{fig:losses}
  	\vspace{-1em}
\end{figure} 

\vspace{-1em}
\paragraph*{Nearest Neighbor (NN) Loss}


For large unlabeled datasets, since we do not have information about $d_i^*$, we cannot compute the loss in Equation~\ref{equ:gt}.  
In lieu of annotated data, we take inspiration from methods such as Iterative Closest Point~\cite{besl1992method} and use the nearest neighbor of our transformed point $\hat{x}_i' = x_i + \hat{d}_i$ as an approximation for the true correspondence. For each transformed point $\hat{x}_i' \in \hat{\mathcal{X}}'$, we find its nearest neighbor $y_j \in \mathcal{Y}$ and compute the Euclidean distance to that point, illustrated as $e_{NN}$ in Figure~\ref{fig:losses}a: 
\begin{equation}
    \mathcal{L}_{NN} = \frac{1}{N}\sum_i^N \min_{y_j \in \mathcal{Y}}\|\hat{x}_i' - y_j\|^2.
\label{equ:nn}
\end{equation}

Assuming that the initial flow estimate is sufficiently close to the correct flow estimate, this loss will bring the transformed point cloud and the target point cloud closer. This loss can, however, have a few drawbacks if imposed alone.  First, the true position of the point $x_i$ transformed by the ground truth flow, $x_i' = x_i + d_i^*$, may not be the same as the position of the nearest neighbor to $\hat{x}'$ (transformed by the \emph{estimated} flow) due to potentially large errors in the estimated flow, as illustrated in Figure~\ref{fig:losses}a.  Further, the position of $x_i'$ may not correspond with any point in $\mathcal{Y}$ if the point cloud $\mathcal{Y}$ is sufficiently sparse, as is common for point clouds collected by sparse 3D LIDAR for autonomous driving.  Finally, this loss does not penalize degenerate solutions where all of the points in $\mathcal{X}$ map to the same point in $\mathcal{Y}$; such a degenerate solution would obtain 0 loss under Equation~\ref{equ:nn}. To address these issues, we use an additional self-supervised loss: cycle consistency loss.

\vspace{-1em}
\paragraph*{Cycle Consistency Loss} 
To avoid the above issues, we incorporate an additional self-supervised loss: cycle consistency loss, illustrated in Figure~\ref{fig:losses}b. We first estimate the ``forward" flow as $\hat{\mathcal{D}} = g(\mathcal{X}, \mathcal{Y})$.  Applying the estimated flow $\hat{d}_i \in \hat{\mathcal{D}}$ to each point $x_i \in \mathcal{X}$ gives an estimate of the location of the point $x_i$ in the next frame: $\hat{x}_i' = x_i + \hat{d}_i$.  We then compute the scene flow in the reverse direction: for each transformed point $\hat{x}_i'$ we estimate the flow to transform the point back to the original frame, $\hat{\mathcal{D'}} = g(\mathcal{\hat{X}'}, \mathcal{X})$.  Transforming each point $\hat{x}_i'$ by this ``reverse" flow $\hat{d}_i'$ gives a new estimated point $\hat{x}_i''$.  If both the forward and reverse flow are accurate, this point $\hat{x}_i''$ should be the same as the original point $x_i$.  The error between these points, $e_{cycle}$, is the ``cycle consistency loss,"  given by
\begin{equation}
    \mathcal{L}_{cycle} = \sum_i^N \|\hat{x}_i'' - x_i\|^2.
\label{equ:cycle}
\end{equation}
A similar loss is used as a regularization in~\cite{liu2019flownet3d}.

%







\begin{figure}[t]
 	\centering
 	\begin{subfigure}{0.23\textwidth}
 	\includegraphics[trim={20 70 20 70}, clip, width=\linewidth]{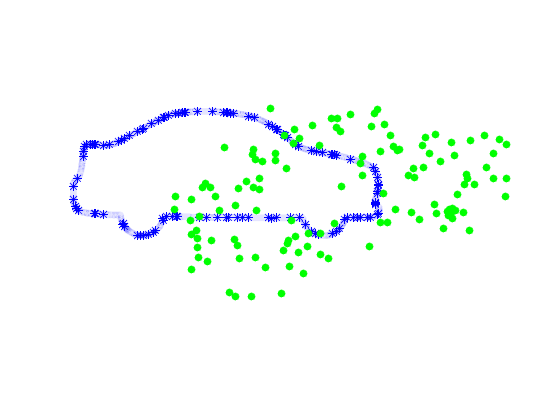}
 	\caption{}
 	\end{subfigure}
 	\hspace*{\fill}
  	\begin{subfigure}{0.23\textwidth}
 	\includegraphics[trim={20 70 20 70}, clip, width=\linewidth]{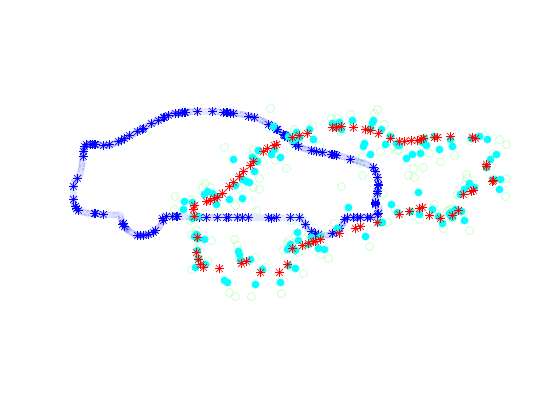}
 	\caption{}
 	\end{subfigure}
 	\caption{Compounding errors cause problems in estimating reverse flow using the transformed point cloud. $(a)$ Large flow prediction errors degrade the structure of the transformed cloud $\hat{\mathcal{X}}'$ (shown in \textcolor{green}{green}). Thus, computing the reverse flow between $\hat{\mathcal{X}}'$ (\textcolor{green}{green}) and $\mathcal{X}$ (\textcolor{blue}{blue}) is an ill-posed task.
 	$(b)$ Using the nearest neighbor points (\textcolor{red}{red}) as anchors, we are able to stabilize the transformed cloud $\bar{\mathcal{X}}'$ (\textcolor{cyan}{cyan}), thus retaining important structural information.}
 	\label{fig:anchor}
  	\vspace{-1em}
\end{figure}

However, we found that implementing the cycle loss in this way can produce unstable results if only self-supervised learning is used without any ground-truth annotations.  These instabilities appear to be caused by errors in the estimated flow which lead to structural distortions in the transformed point cloud $\hat{\mathcal{X}}'$, which is used as the input for computing the reverse flow $g(\mathcal{\hat{X}'}, \mathcal{X})$. This requires the network to simultaneously learn to correct any distortions in $\mathcal{\hat{X}'}$, while also learning to estimate the true reverse flow.
To solve this problem, we use the nearest neighbor $y_j$ of the transformed point $\hat{x}'_i$ as an anchoring point in the reverse pass.  Using the nearest neighbor $y_j$ stabilizes the structure of the transformed cloud while still maintaining the correspondence around the cycle. The effects of this stabilization are illustrated in Figure~\ref{fig:anchor}. As we are using the anchoring point as part of the reverse pass of the cycle, we refer to this loss as ``anchored cycle consistency loss''.


Specifically, we compute the anchored reverse flow as follows. First, we compute the forward flow as before, $\hat{\mathcal{D}} = g(\mathcal{X}, \mathcal{Y})$, which we use to compute the transformed point cloud $\hat{x}_i' = x_i + \hat{d}_i$.  We then compute anchor points $\bar{\mathcal{X}'} = \{\bar{x}_i'\}^N$ as a convex combination of the transformed point and its nearest neighbor $\bar{x}_i' = \lambda \hat{x}_i' + (1-\lambda) y_j$.  In our experiments, we find that $\lambda=0.5$ produces the most accurate results. Finally, we compute the reverse flow using these anchored points: $\bar{\mathcal{D}}' = g(\bar{\mathcal{X}}', \mathcal{X})$.  The cycle loss of Equation~\ref{equ:cycle} is then applied to this anchored reverse flow.  By using anchoring, some of the structural distortion of the predicted point cloud $\mathcal{\hat{X}'}$ will be removed in the anchored point cloud $\bar{\mathcal{X}'}$, leading to a more stable training input for the reverse flow.

Note that the cycle consistency loss also has a degenerate solution: the ``zero flow,", \ie $\hat{\mathcal{D}} = 0$, will produce 0 loss according to Equation~\ref{equ:cycle}.  However, the zero flow will produce produce a non-zero loss when anchored cycle consistency is used; thus anchoring helps to remove this degenerate solution.  Further, the nearest neighbor loss will also be non-zero for the degenerate solution of zero flow.  Thus, the local minima of each of the nearest neighbor and cycle consistency losses conflict, allowing their sum, $\mathcal{L} = \mathcal{L}_{NN} + \mathcal{L}_{cycle}$, to act as a stable surrogate for the true error.


\vspace{-1em}
\paragraph*{Temporal Flip Augmentation}

%
Having a dataset of point cloud sequences in only one direction may generate a motion bias which may lead to the network predicting the flow equal to the average forward speed of the training set. To reduce this bias, we augment the training set by flipping the point clouds, i.e, reversing the flow. With this augmentation, the network sees an equal number of point cloud sequences having a forward motion and a backward motion. 
%


\section{Experiments}

We run several experiments to validate our self-supervised method for scene flow estimation for various levels of supervision and different amounts of data. First, we show that our method, with self-supervised training on large unlabeled datasets, can perform as well as supervised training on the existing labeled data.
Next, we investigate how our results can be improved by combining self-supervised learning with a small amount of supervised learning, exceeding the performance of purely supervised learning. Finally, we explore the utility of each element of our method through an ablation study.



\subsection{Implementation Details}
For all data configurations (our method and the baseline), we initialize our network with the parameters of the Flownet3D model~\cite{liu2019flownet3d} pre-trained on the FlyingThing3D dataset~\cite{mayer2016large}. We compare our self-supervised training procedure to a baseline which uses supervised fine-tuning on the KITTI dataset~\cite{kitti}.  The baseline used in the comparison is the same as in Liu \etal~\cite{liu2019flownet3d}, except that we increase the number of training iterations from 150 epochs (as described in the original paper) to 10k epochs in order to keep the number of training iterations consistent with that used in our self-supervised method. We see that this change leads to a small improvement in the baseline performance, which we include in the results table. 


\begin{figure*}[t]
 	\centering
 	\includegraphics[trim={0 0 0 50}, clip, width=\textwidth]{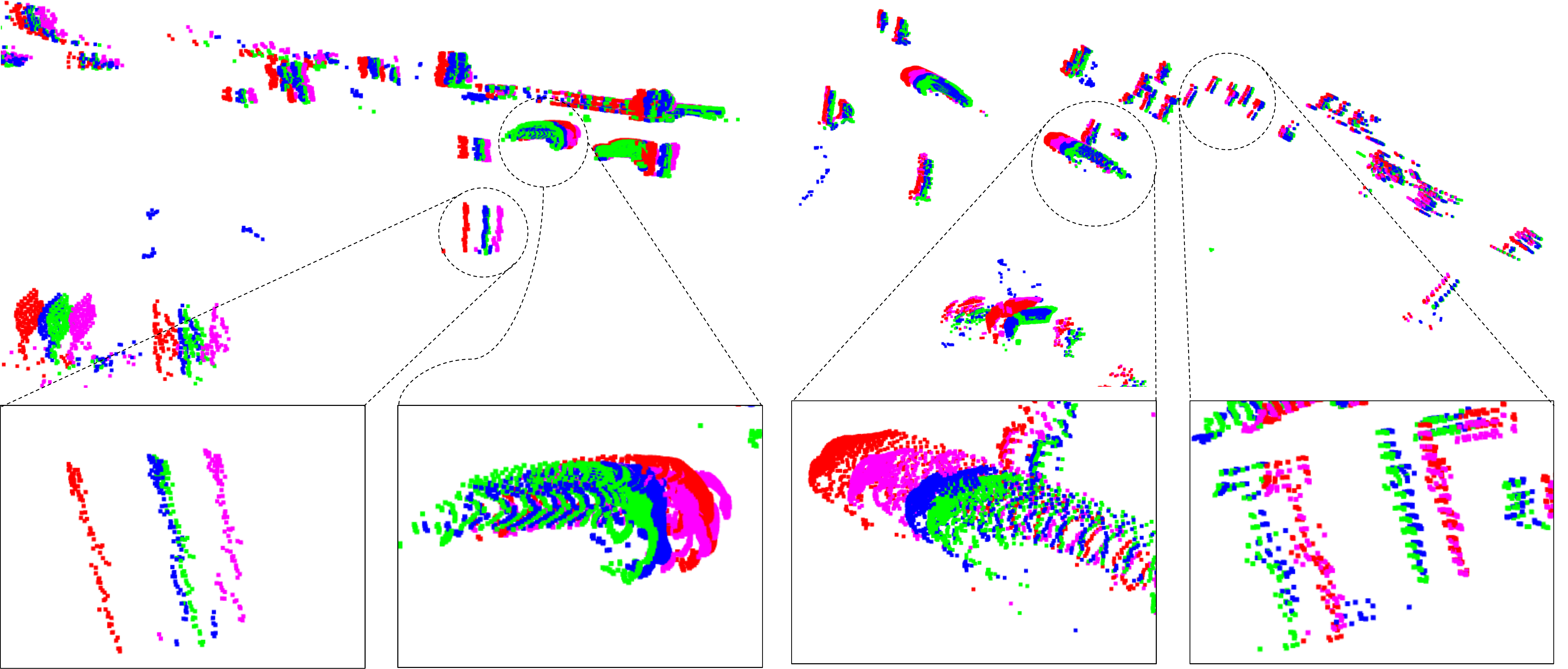}
 	\caption{Scene flow estimation between point clouds at time $t$ (\textcolor{red}{red}) and $t+1$ (\textcolor{green}{green}) from the KITTI dataset trained without any labeled LIDAR data. Predictions from our self-supervised method, trained on nuScenes and fine-tuned on KITTI using self-supervised learning is shown in \textcolor{blue}{blue}; the baseline with only synthetic training is shown in \textcolor{magenta}{purple}. In the absence of real-world supervised training, our method clearly outperforms the baseline method, which overestimate the flow in many regions. (Best viewed in color)}
 	\label{fig:qual_no_ground_truth}
\end{figure*}

\begin{figure*}[t]
 	\centering
 	\begin{subfigure}{0.48\textwidth}
 	\includegraphics[trim={0 0 900 50}, clip, width=\textwidth]{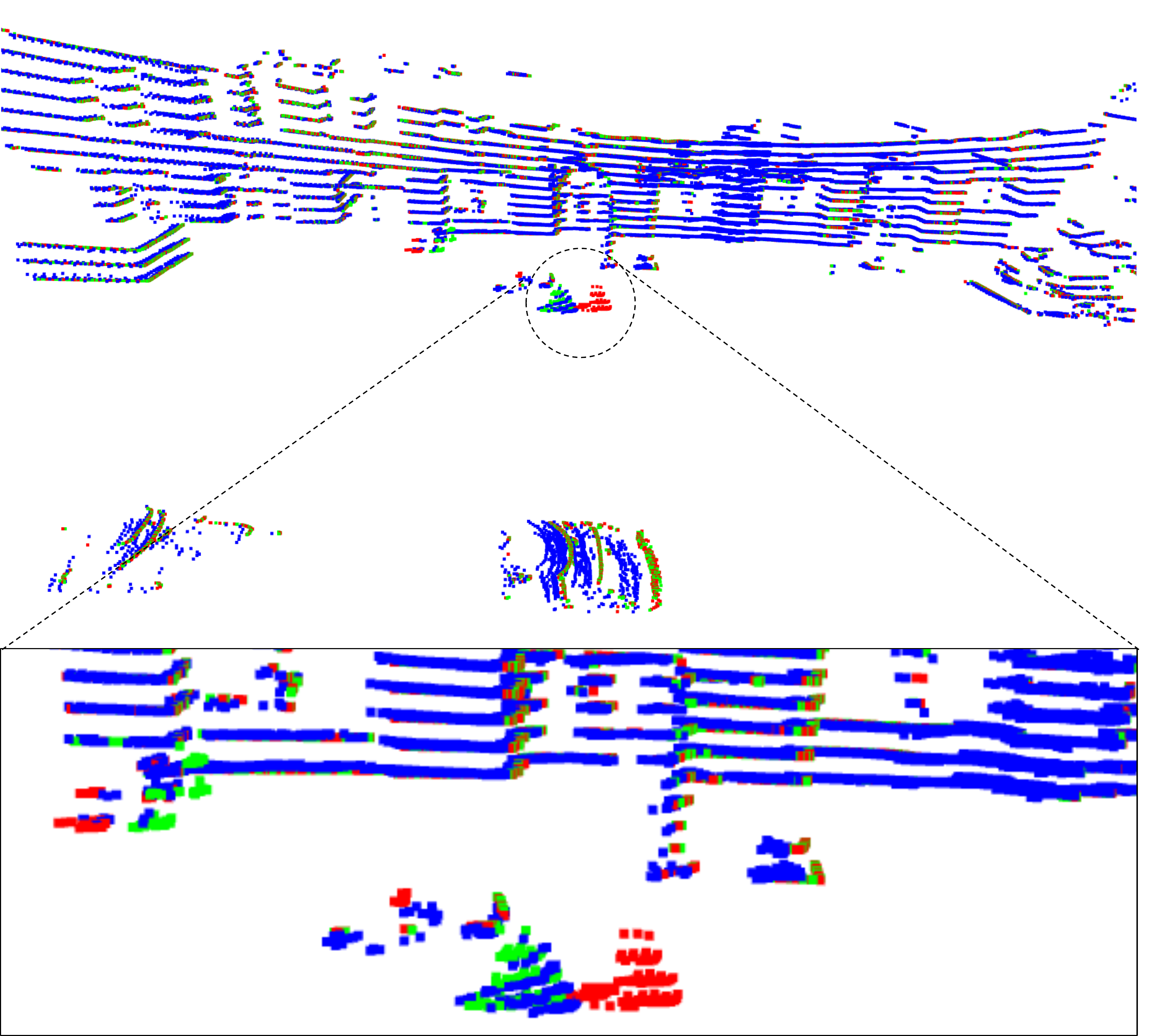}
 	\caption{Ours (Self-Supervised Fine Tuning)}
 	\end{subfigure}
  	\begin{subfigure}{0.48\textwidth}
 	\includegraphics[trim={920 0 0 50}, clip, width=\textwidth]{figures/nuscenes_comparison.pdf}
 	\caption{Baseline (No Fine Tuning)}
 	\end{subfigure}
 	\caption{Comparison of our self-supervised method to a baseline trained only on synthetic data, shown on the nuScenes dataset. Scene flow is computed  between point clouds at time $t$~(\textcolor{red}{red}) and $t+1$~(\textcolor{green}{green}); the point cloud that is transformed using the estimated flow is in shown in \textcolor{blue}{blue}. In our method, the predicted point cloud has a much better overlap with the point cloud of the next timestamp (\textcolor{green}{green})  compared to the baseline. Since nuScenes dataset does not provide any scene flow annotation, the supervised approaches cannot be fined tuned to this environment.}
 	\label{fig:qual_nuscenes}
 	\vspace{-1em}
\end{figure*}

\begin{figure*}[ht!]
 	\centering
 	\includegraphics[trim={0 0 0 40}, clip, width=\textwidth]{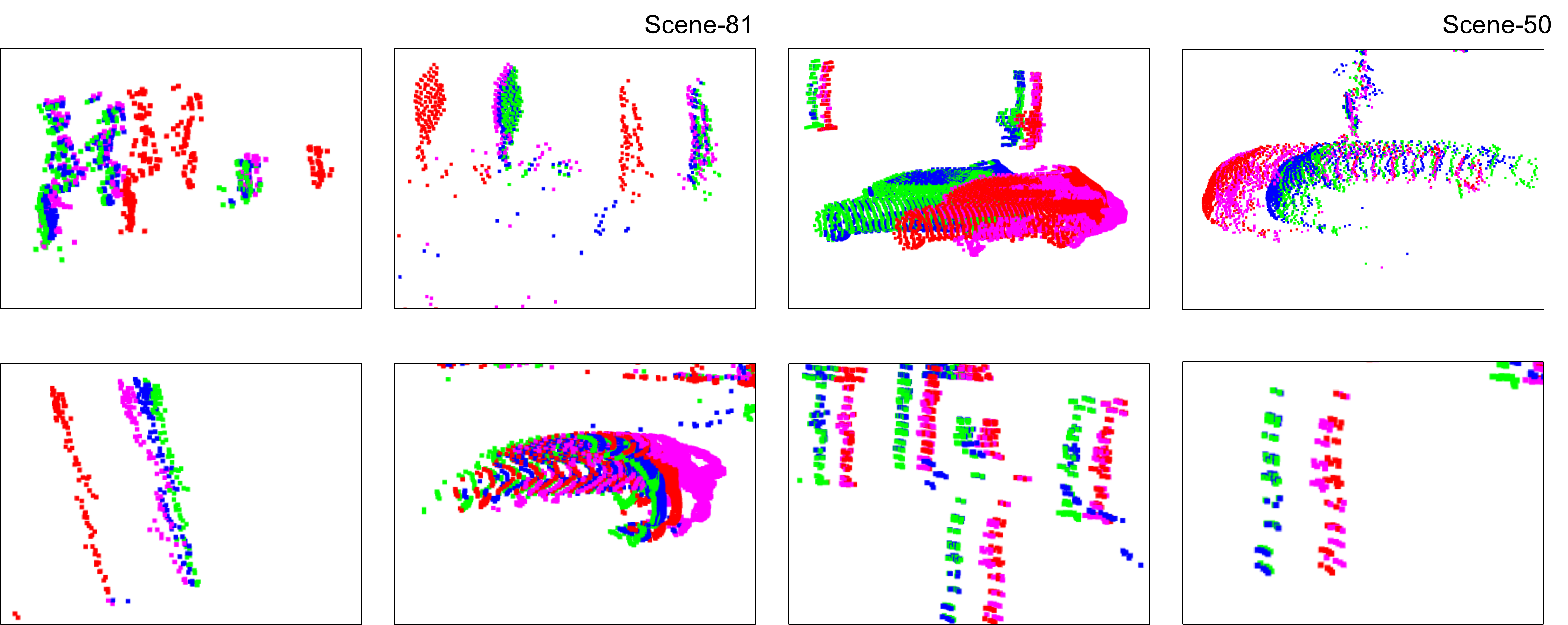}
 	\caption{Scene flow estimation on LIDAR data from the KITTI dataset between point clouds at time $t$ (\textcolor{red}{red}) and $t+1$ (\textcolor{green}{green}). Our method, which is trained on nuScenes using self-supervised learning and then fine-tuned on KITTI using supervised learning, is shown in \textcolor{blue}{blue}. The baseline method is fine-tuned only on KITTI using supervised training and is shown in \textcolor{magenta}{purple}. While in aggregate, both methods well estimate the scene flow, adding self-supervised training on nuScenes (\textcolor{blue}{blue}) enables our predictions to more closely match the next frame point cloud (\textcolor{green}{green}). In several  scenes, the purely supervised method (\textcolor{magenta}{purple}) underestimates the flow, staying too close to the initial point cloud (\textcolor{red}{red}). (Best viewed in color)}
 	\label{fig:qual_supervised}
\end{figure*}

\subsection{Datasets}

\paragraph*{KITTI Vision Benchmark Suite}  KITTI~\cite{kitti} is a real-world self-driving dataset. There are 150 scenes of LIDAR data in KITTI collected using seven scans of a Velodyne 64 LIDAR, augmented using 3D models, and annotated with ground truth scene flow~\cite{menze2018object}. For our experiments under both self-supervised and supervised settings, we consider 100 out of 150 scenes for training and the remaining 50 scenes for testing. Ground points are removed from every scene using the pre-processing that was performed in previous work~ \cite{liu2019flownet3d}. Every scene consists of a pair of point clouds recorded at two different times as well as the ground truth scene flow for every point of the first point cloud. 

\vspace{-1em}
\paragraph*{nuScenes} The nuScenes~\cite{nuscenes2019} dataset is a large-scale public dataset for autonomous driving. It consists of 850 publicly available driving scenes in total from Boston and Singapore. The LIDAR data was collected using a Velodyne 32 LIDAR rotating at 20 Hz. This is in contrast to the 64-beam Velodyne rotating at 10 Hz used for the KITTI dataset.  This difference in sensors leads to a difference in data sparsity that creates a distribution shift between KITTI and nuScenes.  This distribution shift necessitates additional training on KITTI beyond our self-supervised training on nuScenes.  Nonetheless, our results show a substantial benefit from the self-supervised training on nuScenes.

Since the nuScenes dataset~\cite{nuscenes2019} does not contain scene flow annotations, we must use self-supervised methods when working with this dataset. In our experiments, out of the 850 scenes available, we use 700 as the train set and the rest 150 as the validation set. Similar to KITTI, we remove the ground points from each point cloud using a manually tuned height threshold.

\begin{table*}[h]
\centering
\begin{tabular} {l|c c c}
\toprule
 Training Method & EPE (m) $\downarrow$ & ACC (0.05) $\uparrow$ & ACC (0.1) $\uparrow$ \\
\hline
\hline
No Fine Tuning & 0.122 & 25.37\% & 57.85\% \\
KITTI (Supervised) & 
0.100 & 31.42\% & 66.12\% \\
\hline
Ablation: KITTI (Self-Supervised)  & 0.126 & 32.00\% & 73.64\% \\
Ours: nuScenes (Self-Supervised) + KITTI (Self-Supervised)  & 0.105 & 46.48\% & 79.42\% \\
Ours: nuScenes (Self-Supervised) + KITTI (Supervised) & \textbf{0.091} & \textbf{47.92\%} & \textbf{79.63\%} \\
\bottomrule
\end{tabular}
\caption{Comparison of levels of supervision on KITTI dataset. The nearest neighbor + anchored cycle loss is used for nuScenes (self-supervised) and KITTI (self-supervised). All methods are pretrained on FlyingThings3D\cite{mayer2016large} and ground points are removed for KITTI and nuScenes datasets.}\label{tbl:supervision}
\vspace{-1em}
\end{table*}

\subsection{Results}
We use three standard metrics to quantitatively evaluate the predicted scene flow when the ground truth annotations of scene flow are available. Our primary evaluation metric is End Point Error (EPE) which describes the mean Euclidean distance between the predicted and ground truth transformed points, described by Equation~\ref{equ:gt}. We also compute accuracy at two threshold levels, Acc(0.05) as the percentage of scene flow prediction with an EPE~$<~0.05$m or relative error~$<5\%$ and Acc(0.1) as percentage of points having an EPE~$<0.1$m or relative error~$<~10\%$, as was done for evaluation in previous work~\cite{liu2019flownet3d}. 


\subsubsection{Quantitative Results}



\paragraph*{Self-supervised training:}
Unlike previous work, we are not restricted to annotated point cloud datasets; our method can be trained on any sequential point cloud dataset. There are many point cloud datasets containing real LIDAR captures of urban scenes, but most of them do not contain scene flow annotations. Due to lack of annotations, these datasets can not be utilized for supervised scene flow learning. In contrast, our self-supervised loss allows us to easily integrate them into our training set. The combination of these datasets (KITTI + NuScenes) contains 5x more real data than using KITTI alone.


The results are shown in Table~\ref{tbl:supervision}.
To show the value of self-supervised training, we evaluate the performance of our method without using any  ground-truth annotations.  We first pre-train on the synthetic FlyingThings3D dataset; we then perform self-supervised fine-tuning on the large nuScenes dataset followed by further self-supervised fine-tuning on the smaller KITTI dataset (4th row: ``Ours: nuScenes (Self-Supervised) + KITTI (Self-Supervised)").  As can be seen, 
using no real-world annotations, we are able to achieve an EPE of 0.105~m.  This outperforms the baseline of only training on synthetic data (``No Fine Tuning").  Even more impressively, our approach performs similarly to the baseline which 
pre-trains on synthetic data and then 
does supervised fine-tuning  on the KITTI dataset (``KITTI (Supervised)"); our method has a similar EPE and outperforms this baseline in terms of accuracy, despite not having access to any annotated training data. This result shows that our method for self-supervised training, with a large enough unlabeled dataset, can match the performance of supervised training.

\vspace{-1em}
\paragraph*{Self-supervised + Supervised:} Finally, we show the value of combining our self-supervised learning method with a small amount of supervised learning. For this analysis, we perform self-supervised training on NuScenes as above, followed by supervised training on the much smaller KITTI dataset.  The results are shown in the last row of Table~\ref{tbl:supervision}.

As can be seen, this approach of self-supervised training followed by supervised fine-tuning outperforms all other methods on this benchmark, obtaining an EPE of 0.091, outperforming the previous state-of-the-art result which used only supervised training.  This shows the benefit of self-supervised training on large unlabeled datasets to improve scene flow accuracy, even when scene flow annotations are available.

While Table~\ref{tbl:supervision} only shows results using the FlowNet3D~\cite{liu2019flownet3d} architecture, we note that our method also outperforms the results of HPLFlownet~\cite{gu2019hplflownet}~(EPE of 0.1169) and all models they compare against as well.

\begin{figure}[t]
 	\centering
 	\includegraphics[width=0.47\textwidth]{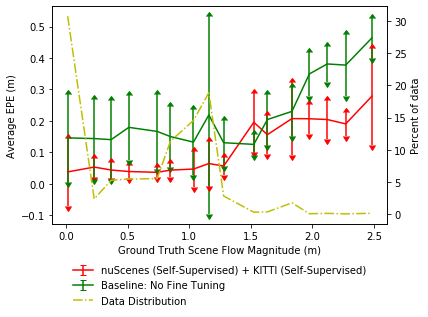}
 	\caption{Analysis of average EPE (m) with respect to ground truth flow magnitudes (m). Flow estimates are binned by ground truth flow magnitude and a confidence interval of 95\% is shown for all results.}
 	\label{fig:epe_vs_mag}
 	\vspace{-1em}
\end{figure}

Figure \ref{fig:epe_vs_mag} provides an analysis on the correlation between average endpoint error and the magnitude of the ground truth flow.  As can be seen, our method consistently outperforms the baseline at almost all flow magnitudes. 


\subsubsection{Qualitative Analysis}
\paragraph*{Self-supervised training - KITTI results:} 
Next we perform a qualitative analysis to visualize the performance of our method.
We compare our method (synthetic pre-training + self-supervised training on nuScenes + self-supervised training on KITTI) compared to the baseline of synthetic training only.  Results on KITTI are shown in Figure~\ref{fig:qual_no_ground_truth}. The figure shows the point clouds captured at time $t$ and $t+1$  in red and green, respectively. The predictions from our method are shown in blue and the baseline predictions are shown in purple. As shown, our scene flow predictions (blue)  have a large overlap with the point cloud at time $t+1$ (green). On the other hand, the baseline predictions (purple) do not overlap with the point cloud at time $t+1$.  The baseline, trained only on synthetic data, fails to generalize to the real-world KITTI dataset. On the contrary, our self-supervised approach can be fine tuned on any real world environment and shows a significant improvement over the baseline. 



\vspace{-1em}
\paragraph*{Self-supervised training - nuScenes results:}



Next, we visualize the performance of our method on the nuScenes dataset.  Note that, because nuScenes does not have scene flow annotations, only qualitative results can be shown on this dataset. For this analysis, our method is pre-trained on synthetic data (FlyingThings3D) as before and then fine-tuned on nuScenes in a self-supervised manner. No scene flow annotations for nuScenes are available, so we compare to a baseline which is trained only on synthetic data.  

The results on nuScenes are shown in Figure~\ref{fig:qual_nuscenes}. These results again showcase the advantages of self-supervision on real world data over purely synthetic supervised training. As before, the figure shows the point clouds captured at time $t$ and $t+1$ in red and green respectively. The predictions are shown in blue, with our method on the left (Fig.~\ref{fig:qual_nuscenes}a) and the baseline on the right (Fig.~\ref{fig:qual_nuscenes}b).  As shown, our scene flow predictions (left figure, blue) have a large overlap with the point cloud at time $t+1$ (green). On the other hand, the baseline predictions (right figure, blue) do not overlap with the point cloud at time $t+1$. 



The low performance of the baseline can again be attributed to its training on synthetic data and its inability to generalize to real-world data.  For the nuScenes data, no scene flow annotations exist, so only self-supervised learning is feasible to improve performance.


\begin{table}[]
\centering
\footnotesize
\setlength\tabcolsep{2pt}
\begin{tabular}{c|c|c|c||c|c|c}
\toprule
\multicolumn{7}{l}{Self-Supervised Training (nuScenes + KITTI)} \\ \midrule
NN Loss & Cycle Loss & Anchor Pts & Flip & EPE (m) & ACC (0.05) & ACC (0.1) \\ 
\hline
\checkmark & \checkmark & \checkmark & \checkmark & 0.105 & 46.48\% & 79.42\% \\ 
\checkmark & \checkmark & \checkmark &  & 0.107 & 40.03\% & 72.20\% \\
\checkmark & \checkmark &  &  & 0.146 & 30.21\% & 48.57\% \\
\checkmark &  &  &  & 0.108 & 42.00\% & 78.51\% \\
 &  &  &  & 0.122 & 25.37\% & 57.85\% \\
\hline \hline
\multicolumn{7}{l}{Self-Supervised (nuScenes) + Supervised Training (KITTI)} \\ \hline
NN Loss & Cycle Loss & Anchor Pts & Flip & EPE (m) & ACC (0.05) & ACC (0.1) \\ 
\hline
\checkmark & \checkmark & \checkmark & \checkmark & {0.091} & {47.92\%} & {79.63\%} \\ 
\checkmark & \checkmark & \checkmark &  & {0.093} & {40.69\%} & {74.50\%} \\
\checkmark & \checkmark &  &  & 0.092 & 30.76\% & 72.94\% \\
\checkmark &  &  &  & 0.114 & 31.24\% & 64.58\% \\
 &  &  &  & 0.100 & 31.42\% & 66.12\%\\ \bottomrule
\end{tabular}
\caption{Ablation analysis: We study the effect of the different self-supervised losses and data augmentation. Top: Models use self-supervised training on nuScenes and KITTI; Bottom: Models use self-supervised training on nuScenes followed by supervised training on KITTI.}\label{tbl:ablation}
\vspace{-1em}
\end{table}



\begin{table}
\centering
\footnotesize
\renewcommand{\arraystretch}{1.2}
\begin{tabular}{c | c | c | c}
\toprule
$\lambda$ & EPE (m)$\downarrow$ & ACC (0.05)$\uparrow$ & ACC (0.1)$\uparrow$ \\ 
\hline
0   & 0.120 & 24.09\% & 73.20\% \\
0.25 & 0.122 &  26.41\% & 65.57\% \\
0.5 & \textbf{0.105} & \textbf{46.48\%} & \textbf{79.42\%} \\
0.75 &  0.125 & 23.59\% & 62.96\% \\
1   & 0.149 & 22.97\% & 49.58\% \\
\bottomrule
\end{tabular}
\caption{Effect of varying the $\lambda$ parameter for ``anchoring" the Cycle Consistency Loss. Results are shown for self-supervised training on nuScenes + KITTI.}\label{tbl:cycle_lambda}
\vspace{-1em}
\end{table}


\vspace{-1em}
\paragraph*{Self-supervised + supervised training - KITTI results:} 
Finally, we show the value of combining our self-supervised learning method with a small amount of supervised learning, compared to only performing supervised learning.  For our method, we perform synthetic pre-training, followed by self-supervised fine-tuning on nuScenes, followed by supervised fine-tuning on the much smaller KITTI dataset. We compare to a baseline which only performs synthetic pre-training followed by supervised fine-tuning KITTI.




Qualitative results can be seen in Figure~\ref{fig:qual_supervised}. The figure shows the point clouds captured at time $t$ and $t+1$  in red and green, respectively. The predictions from our method are shown in blue and the baseline predictions are shown in purple. As shown, our scene flow predictions (blue)  have a large overlap with the point cloud at time $t+1$ (green), whereas the baseline predictions (purple) do not. 


The baseline predicts a small motion, keeping the transformed cloud (purple) too close to the initial position (red). As discussed above, this bias towards small motion is likely due to the training of the baseline over a synthetic dataset, which affects the generalization of the baseline to real-world datasets where objects exhibit different types of motion than seen in simulation. By training over a significantly larger unlabeled dataset, our method is able to avoid overfitting and generalizes better to the scenes and flow patterns which were not present in the synthetic dataset.

\begin{figure}[t]
 	\centering
 	\includegraphics[trim={0 0 0 100}, clip, width=\columnwidth]{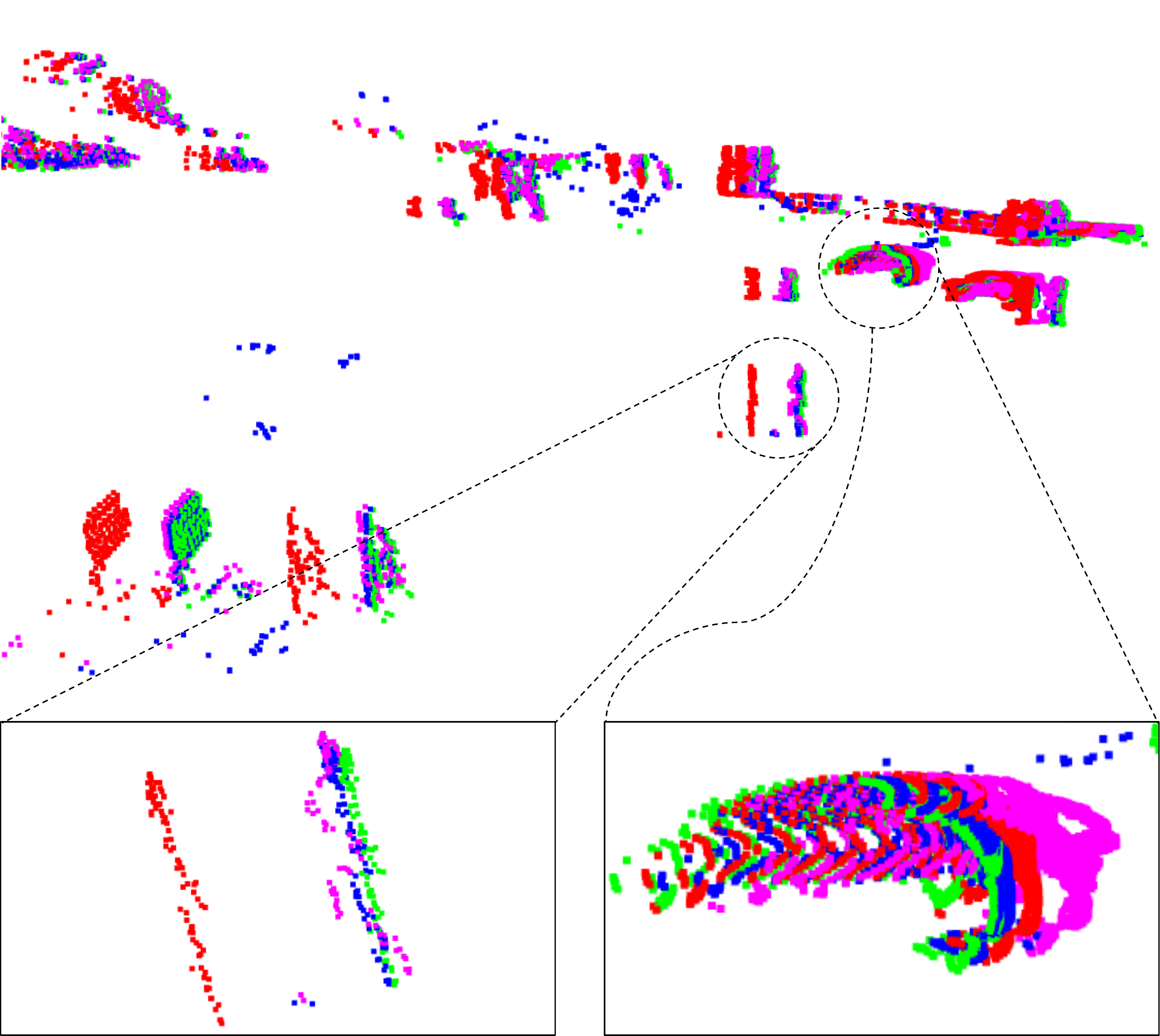}
 	\caption{Ablation study comparing our self-supervised method with both nearest neighbor loss and anchored cycle consistency loss (\textcolor{blue}{blue}) compared to training only using the nearest neighbor loss (\textcolor{magenta}{purple}). Scene flow is computed between point clouds from the KITTI dataset at time $t$ (\textcolor{red}{red}) and $t+1$ (\textcolor{green}{green})}
 	\label{fig:nuscenes_ablation}
  	\vspace{-1em}
\end{figure}

\subsubsection{Ablation Studies}


We test the importance of each component of our method by running a series of ablation studies. Table~\ref{tbl:ablation} shows the effect of iteratively removing portions of our method in both the purely self-supervised (top) and self-supervised + supervised (bottom) training.
The benefits of anchored cycle consistency loss (compared to only using nearest neighbor loss) can be seen in  Table~\ref{tbl:ablation} (bottom) as well as in Figure~\ref{fig:nuscenes_ablation}. 

The benefits of anchoring is apparent by the large drop in performance for self-supervised training (top) when anchoring is removed.  
 Introducing the anchor point cloud as part of the backward flow greatly improves performance when only self-supervised training is used (Figure~\ref{fig:nuscenes_ablation}, top).
  This suggests that having anchored point clouds stabilizes the training of the cycle consistency loss. Additionally, we evaluate the sensitivity of our system to the selection of the anchoring parameter, $\lambda$. Table~\ref{tbl:cycle_lambda} shows that we obtain the best results with $\lambda=0.5$, \ie using the average of the predicted and nearest points. Overall, these analyses show the benefits of each component of our method. Further ablation analysis can be found in the supplement.
 

\section{Conclusion}


In this work, we propose a self-supervised method for training scene flow algorithms using a combination of cycle consistency in time and nearest neighbor losses. Our purely self-supervised method is able to achieve a performance comparable to that of the supervised methods on the widely used KITTI self-driving dataset. We further show that when supervised training is augmented with self-supervision on a large-scale, unannotated dataset, the results exceed the current state-of-the-art performance. Our self-supervision method opens the door to fine-tuning on arbitrary datasets that lack scene flow annotations. 


\subsection*{Acknowledgements}
This work was supported by the CMU Argo AI Center for Autonomous Vehicle Research and a NASA Space Technology Research Fellowship.

{\small
\bibliographystyle{ieee_fullname}
\bibliography{egbib}
}
\newpage
\section*{Supplementary}
\subsection{Hyperparameters}
A batch size of 8 is used for all datasets. All models are trained using the Adam optimizer with a learning rate of 1e-4, $\beta_1=0.9$, $\beta_2=0.999$, and $\epsilon=1e-8$.

\subsection{Further Ablation Studies}
In this section, we extend our ablation study to further evaluate the importance of each component used in our method. Beginning from our full method, we remove a single component to see the change on the evaluation metrics.

The results can be seen in Table~\ref{tbl:ablation}.  Because  ``anchoring" is a modification of the cycle loss, when Cycle Consistency Loss is removed, anchoring must be removed too.  The results in Table~\ref{tbl:ablation} show that each component is important for the performance of our method, and removing any of the components results in a drop in performance. This can especially be seen in the accuracy performance metrics (ACC 0.05 and 0.1), which show large drops when any component is removed.

\begin{table}[h]
\centering
\footnotesize
\setlength\tabcolsep{2pt}
\begin{tabular}{c|c|c|c||c|c|c}
\toprule
\multicolumn{7}{l}{Self-Supervised Training (nuScenes + KITTI)} \\ \midrule
NN Loss & Cycle Loss & Anchor  & Flip & EPE (m)$\downarrow$ & ACC (0.05)$\uparrow$ & ACC (0.1)$\uparrow$ \\ 
\hline
           & \checkmark & \checkmark & \checkmark & 0.1768 & 15.90 & 35.81 \\
\checkmark &            &            & \checkmark & 0.1102 & 30.80 & 73.27 \\ 
\checkmark & \checkmark &  & \checkmark & 0.1493 & 22.97 & 49.58 \\
\checkmark & \checkmark &  \checkmark &  & 0.1072 & 40.03 & 72.20 \\
\hline
\checkmark & \checkmark & \checkmark & \checkmark & \textbf{0.1053} & \textbf{46.48} & \textbf{79.42} \\
\hline \hline
\multicolumn{7}{l}{Self-supervised (nuScenes) + Supervised Training (KITTI)} \\ \hline
NN Loss & Cycle Loss & Anchor & Flip & EPE (m)$\downarrow$ & ACC (0.05)$\uparrow$ & ACC (0.1)$\uparrow$ \\ 
\hline
          & \checkmark & \checkmark & \checkmark & {0.1572} & {18.50} & {52.80} \\
\checkmark &            &            & \checkmark & {0.1090} & {34.88} & {71.32} \\ 
\checkmark & \checkmark &  & \checkmark & {0.0932} & {28.18} & {66.10} \\
\checkmark & \checkmark &  \checkmark &  & 0.0926 & 40.69 & 74.50 \\
\hline
\checkmark & \checkmark & \checkmark & \checkmark & \textbf{0.0912} & \textbf{47.92} & \textbf{79.63}  \\
\bottomrule
\end{tabular}
\caption{Leave-one-out ablation analysis: We study the effect of removing a single component of self-supervised loss and data augmentation. Top: Models use self-supervised training on nuScenes and KITTI; Bottom: Models use self-supervised training on nuScenes followed by supervised training on KITTI.}\label{tbl:ablation}
\vspace{-1em}
\end{table}

\subsection{Analysis of Point Density vs End Point Error}

\begin{figure}
  \includegraphics[width=\linewidth]{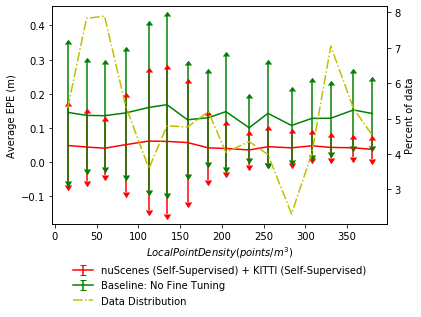}
  \caption{Analysis of average EPE (m) with respect to the local point density (points/$m^3$). Flow estimates are binned by local point density and a confidence interval of 95\% is shown for all results.}
  \label{fig:query_ball}
\end{figure}

We analyze the correlation between local point density and endpoint error. For each point in the point cloud at time $t$, we compute the number of points within a 0.1m radius neighborhood of that point. Each point is binned based on the density of its local neighborhood. For each bin, the mean end-point error (EPE) is computed for both the baseline, trained only synthetic data, and our method, trained on nuScenes and KITTI using our self-supervised losses. Figure \ref{fig:query_ball} shows no correlation between EPE and neighborhood density for either the baseline or for our method.

\subsection{Error Distribution of End Point Error}

\begin{figure}
  \includegraphics[width=\linewidth]{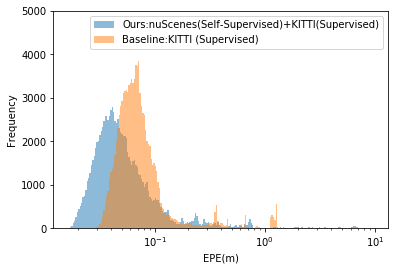}
  \caption{End point error distribution of our method (\textcolor{blue}{blue}) and the baseline (\textcolor{orange}{orange}).}
  \label{fig:error_distribution}
\end{figure}

We analyze the error distributions of our method and a baseline method. Our method uses self-supervised training on nuScenes followed by supervised training on KITTI. The baseline is trained only on KITTI using supervised learning. By computing the error at every point of every scan in the KITTI test set, we can view the full distribution of end point errors (EPE), shown in Figure~\ref{fig:error_distribution}. To better show the effects of outliers, we use log binning for the x-axis. Not only is the center of our error distribution lower, with an average EPE of 0.091, but it also shows fewer large outliers than the baseline method. 

\end{document}